\documentclass[letterpaper, 10 pt, conference]{ieeeconf}  
\IEEEoverridecommandlockouts                              
\usepackage{graphics} 
\usepackage{epsfig} 
\usepackage{mathptmx} 
\usepackage{times} 
\usepackage{amsmath} 
\usepackage{amssymb}  
\usepackage{amsfonts}       

\usepackage{bbding}
\usepackage[utf8]{inputenc} 
\usepackage[T1]{fontenc}    
\usepackage{hyperref}       
\usepackage{url}            
\usepackage{booktabs}       

\usepackage{nicefrac}       
\usepackage{microtype}      
\usepackage{xcolor}         
\usepackage{cleveref}
\usepackage{graphicx}
\usepackage{float}
\usepackage{diagbox}

\usepackage{algorithm}
\usepackage{algpseudocode}
\usepackage{arydshln}
\usepackage{multirow}
\usepackage{listings}
\usepackage{ulem}

\definecolor{dkgreen}{rgb}{0,0.6,0}
\definecolor{gray}{rgb}{0.5,0.5,0.5}
\definecolor{mauve}{rgb}{0.58,0,0.82}
\title{\LARGE \bf
Bridging Zero-shot Object Navigation and Foundation Models through 
Pixel-Guided Navigation Skill
}

\author{Wenzhe Cai, Siyuan Huang, Guangran Cheng, Yuxing Long, Peng Gao, Changyin Sun, Hao Dong 
\thanks{
Wenzhe Cai, Guangran Cheng and Changyin Sun are with the School of Automation, Southeast University.
Yuxing Long and Hao Dong are with Hyperlane Lab, CFCS, School of CS, Peking University and National Key Laboratory for Multimedia Information Processing.
Siyuan Huang and Peng Gao are with the Shanghai AI Laboratory.
}
}  

\begin{document}
\maketitle
\begin{abstract} Zero-shot object navigation is a challenging task for home-assistance robots. This task emphasizes visual grounding, commonsense inference and locomotion abilities, where the first two are inherent in foundation models. But for the locomotion part, most works still depend on map-based planning approaches. The gap between RGB space and map space makes it difficult to directly transfer the knowledge from foundation models to navigation tasks. In this work, we propose a Pixel-guided Navigation skill~(PixNav), which bridges the gap between the foundation models and the embodied navigation task. It is straightforward for recent foundation models to indicate an object by pixels, and with pixels as the goal specification, our method becomes a versatile navigation policy towards all different kinds of objects. Besides, our PixNav is a pure RGB-based policy that can reduce the cost of home-assistance robots. Experiments demonstrate the robustness of the PixNav which achieves 80+\% success rate in the local path-planning task. To perform long-horizon object navigation, we design an LLM-based planner to utilize the commonsense knowledge between objects and rooms to select the best waypoint. Evaluations across both photorealistic indoor simulators and real-world environments validate the effectiveness of our proposed navigation strategy. Code and video demos are available at \href{https://github.com/wzcai99/Pixel-Navigator}{https://github.com/wzcai99/Pixel-Navigator}.
\end{abstract}

\section{INTRODUCTION}
Zero-shot object navigation is considered fundamental for home-assistance robots. Before a robot can interact with objects and complete many complex tasks, the robot should be able to navigate and find the target objects with a high success rate. Besides, with all kinds of different human demands, the robot navigation policy should be able to generalize to a variety of objects. This task requires three essential abilities: visual grounding for novel objects recognition, commonsense inference for semantic exploration and path-planning for robot locomotion. For the first two necessities, recent prosperity in foundation models provide powerful visual and language perception capabilities~\cite{Liu2023GroundingDM,Minderer2022SimpleOO,kirillov2023segment,Li2022LanguagedrivenSS,zhang2023llama,Dai2023InstructBLIPTG}. But for the locomotion part, most of the zero-shot object navigation works still tethered to the map-based planning approaches~\cite{Gadre2022CLIPOW,zhou2023esc,Yu2023L3MVNLL,dorbala2023can,chen2023train}. There are two main limitations of such navigation systems. Firstly, as most foundation models are trained with internet RGB images, it is difficult to directly integrate such prior knowledge into a map-based locomotion method. Secondly, map-based approaches require accurate depth sensing and localization, which often comes at a higher cost and can be easily disturbed compared with RGB cameras. This raises a pertinent question: Can we utilize the superpower of foundation models and develop a pure RGB-based intelligent navigation system? 

\begin{figure}[t]
\begin{center}
    \includegraphics[width=\linewidth]{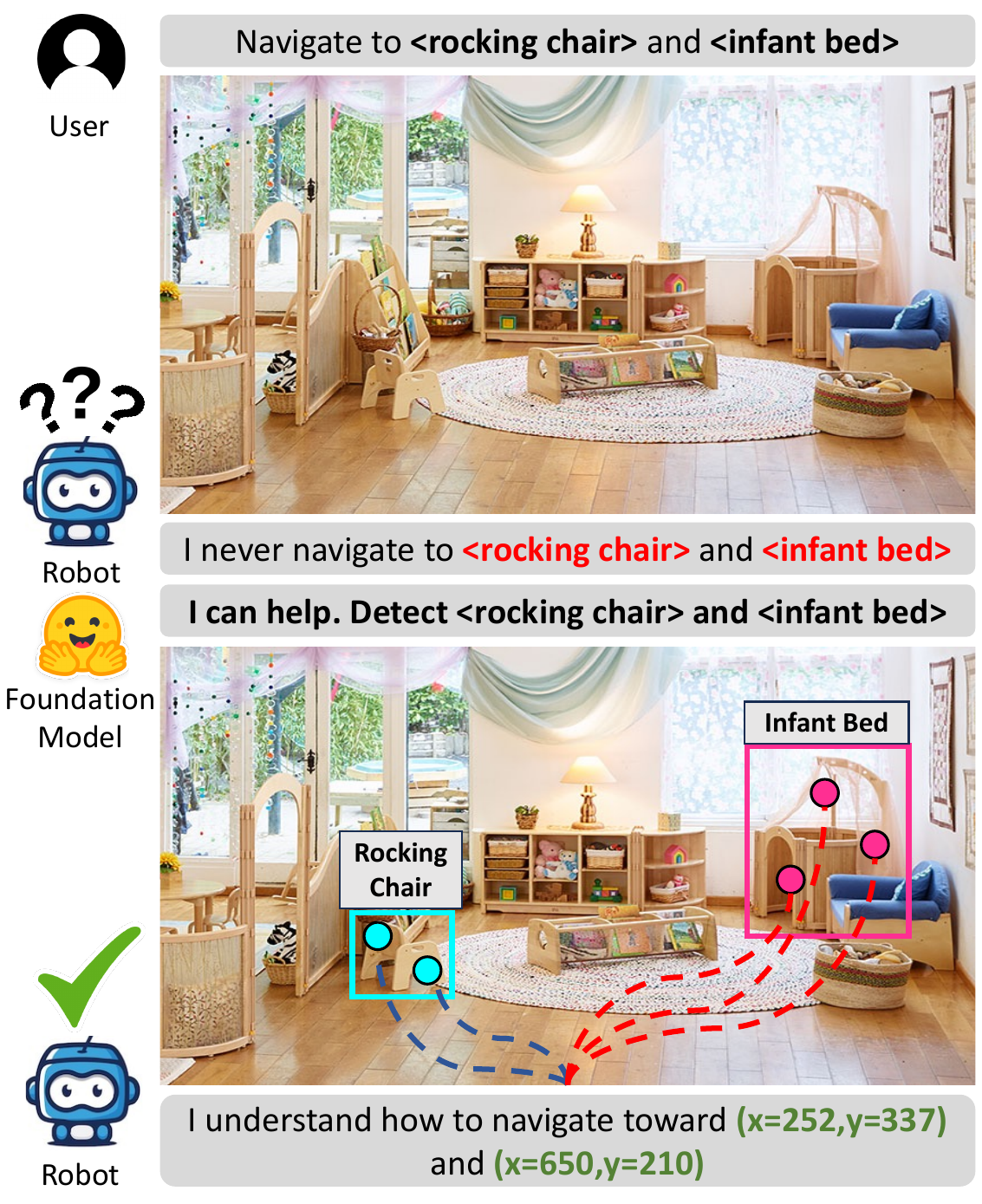}
    \vspace{-0.9cm}
\end{center}
\caption{For the home-assistance robot, humans may ask for searching and interacting with uncommon objects like `a rocking chair' and `a infant bed'. It is challenging for the robot to develop a multi-modal navigation policy that can accommodate various types of objects using both text and image inputs. However, since foundation models are capable of zero-shot image understanding, any objects can be indicated by a single pixel when utilizing such foundation models. Therefore, we train a navigation policy with pixels as goal specifications, enabling the robot to navigate towards arbitrary objects. Since each object typically comprises hundreds of pixels, this approach offers a wide range of possible navigation trajectories, which greatly enlarges the scale of the navigation dataset.}
\label{fig:teasor}
\vspace{-0.6cm}
\end{figure}

In this work, we posit that pixels serve as an ideal bridge, seamlessly connecting foundational models to the navigation task. Since most foundation models are trained with RGB images, aligning such foundation models with an RGB-based navigation skill is much easier than a map-based skill. Meanwhile, the task of navigating towards arbitrary objects can be converted into navigating towards a designated pixel. Therefore, we propose a \textbf{Pix}el-guided \textbf{Nav}igation skill~(PixNav) to complete the last-mile navigation process. Different from the previous object-goal~\cite{khandelwal2022:embodied-clip,chaplot2020object} skill, which uses a word or one-hot vector as the goal specification, PixNav avoids learning the multi-modal alignment for the specific navigation task but still versatile for navigating to any kind of objects. Moreover, to ensure the generalization performance of a learning-based robotics policy, the dataset scale is a crucial factor \cite{Brohan2022RT1RT, Walke2023BridgeDataVA}. For our PixNav, it is much easier to collect large-scale demonstrations than vanilla object navigation. For example, in Fig~\ref{fig:teasor}, we can only get one trajectory for the task "navigating towards infant bed" with an explicit focus on "infant bed" in the object navigation setting. However, we can gather multiple navigation trajectories for PixNav by assigning different pixels belonging to the "infant bed" as the goal. Given that each image comprises thousands of pixels, the potential for generating diverse trajectories is vast. Compared with object navigation, gathering PixNav trajectories doesn't need explicit semantic annotations, which further reduces the cost of data collection. Generating the optimal actions towards the pixel is also simple: Each pixel is an underlying 3-D coordinates and we can use off-the-shelf planning methods to achieve this point-goal navigation for dataset collection.

The remaining question of PixNav is how to navigate towards the out-of-sight target. For indoor navigation scenarios, the house's layout and object placements typically adhere to human preferences, and such commonsense priors have already been embedded into large-language models~(LLMs). Therefore, we designed a hierarchical policy that employs LLM as the planner to guide the exploration sequences across rooms. To make the planning consistent and rational, we designed a step-by-step prompt template to evoke the reasoning ability of LLM. With our designed prompt, the LLM demonstrates its potential in self-localization and structural memory organization. Our hierarchical policy achieves competitive performance compared with the mapping-based methods in zero-shot object navigation without depth modality as input. In summary, our contributions are listed below:
\begin{itemize}
    \item \noindent \textbf{Pixel Navigation} -  We propose a universal RGB-based navigation policy to replace the function of the path-planning method in map-based navigation. 
    \item \noindent  \textbf{Chain of Thought of LLM for Navigation} - We design a step-by-step prompt template and discover the potential of LLM in self-localization and structural memory organization.
    \item \noindent \textbf{Foundation Model Powered Navigation} - Our approach bridges the gap between the embodied navigation tasks and the foundation models. Our hierarchical policy achieves competitive performance compared with the mapping-based methods.
\end{itemize}

\section{RELATED WORK}

\subsection{Open-Vocabulary Object Navigation}
Searching for an object is a common task in indoor environments. While there is a rich body of work dedicated to the object navigation task~\cite{chaplot2020object,rramrakhya2022,ramrakhya2023pirlnav,yadav2023offline,khandelwal2022:embodied-clip,Zhu2022NavigatingTO}, most are limited in a close-set of objects. In response, many researchers study the more realistic open-vocabulary object navigation task. ZSON \cite{majumdar2022zson} transfers an image-goal navigation policy for object-goal navigation tasks with the text-image alignment from the CLIP \cite{Radford2021LearningTV} encoder. 
And SSNet \cite{Zhao2022ZeroShotOG} explicitly introduces an encoding network to extract the feature for the target object tags. Without a large-scale object navigation dataset, the end-to-end policy network suffers in learning the multi-modal alignment for such embodied task and achieved limit performance. Therefore, many map-based approaches \cite{Gadre2022CLIPOW,Huang2022VisualLM,Chen2022OpenvocabularyQS,Yu2023L3MVNLL} are proposed where a high-level policy selects a promising coordinate for object searching and the low-level policy complete the point-goal navigation with path-planning. Yet, the map-based approaches are highly dependent on the accuracy of depth-sensor and self-localization. On the contrary, we propose a fundamental pixel navigation policy that can replace the traditional path-planning method and make the approach become a pure RGB-based solution.

\subsection{Foundation Models on Vision and Language}
Recent advancements have seen significant strides in foundational models across both visual and language domains. For example,  GroundingDINO~\cite{Liu2023GroundingDM} and Owl-ViT~\cite{Minderer2022SimpleOO} can accomplish zero-shot detection, RAM~\cite{Zhang2023RecognizeAA} can tag almost all the appeared objects in the image. SAM~\cite{kirillov2023segment} pioneers in delivering fine-granular segmentation for a diverse set of objects. As for language, LLM shows surprising performance on almost all the tasks, including text understanding, generation~\cite{Touvron2023LLaMAOA, Brown2020LanguageMA} and even planning for robotics field~\cite{Huang2022LanguageMA, 10161317,rana2023sayplan, Huang2023Instruct2ActMM}. However, there are many tasks that require both vision and language inputs. To this end, many works introduce the large-language model as the backbone and develop vision-language models for general-purpose multi-modal tasks ~\cite{Zhu2023MiniGPT4EV, Dai2023InstructBLIPTG, Zhang2023LLaMAAdapterEF}. 
Within the robotics arena, PALM-E~\cite{Driess2023PaLMEAE} is a specialized foundation model that grounds the images and language into several embodied tasks. In our work, we utilize the LLama-Adapter~\cite{Zhang2023LLaMAAdapterEF} to convert images into text and employ GPT-4 to facilitate reasonable textual planning. Furthermore, leverage the robust zero-shot recognition capabilities of foundational models to enhance performance.

\begin{figure*}[t]
\begin{center}
    \includegraphics[width=0.90\linewidth]{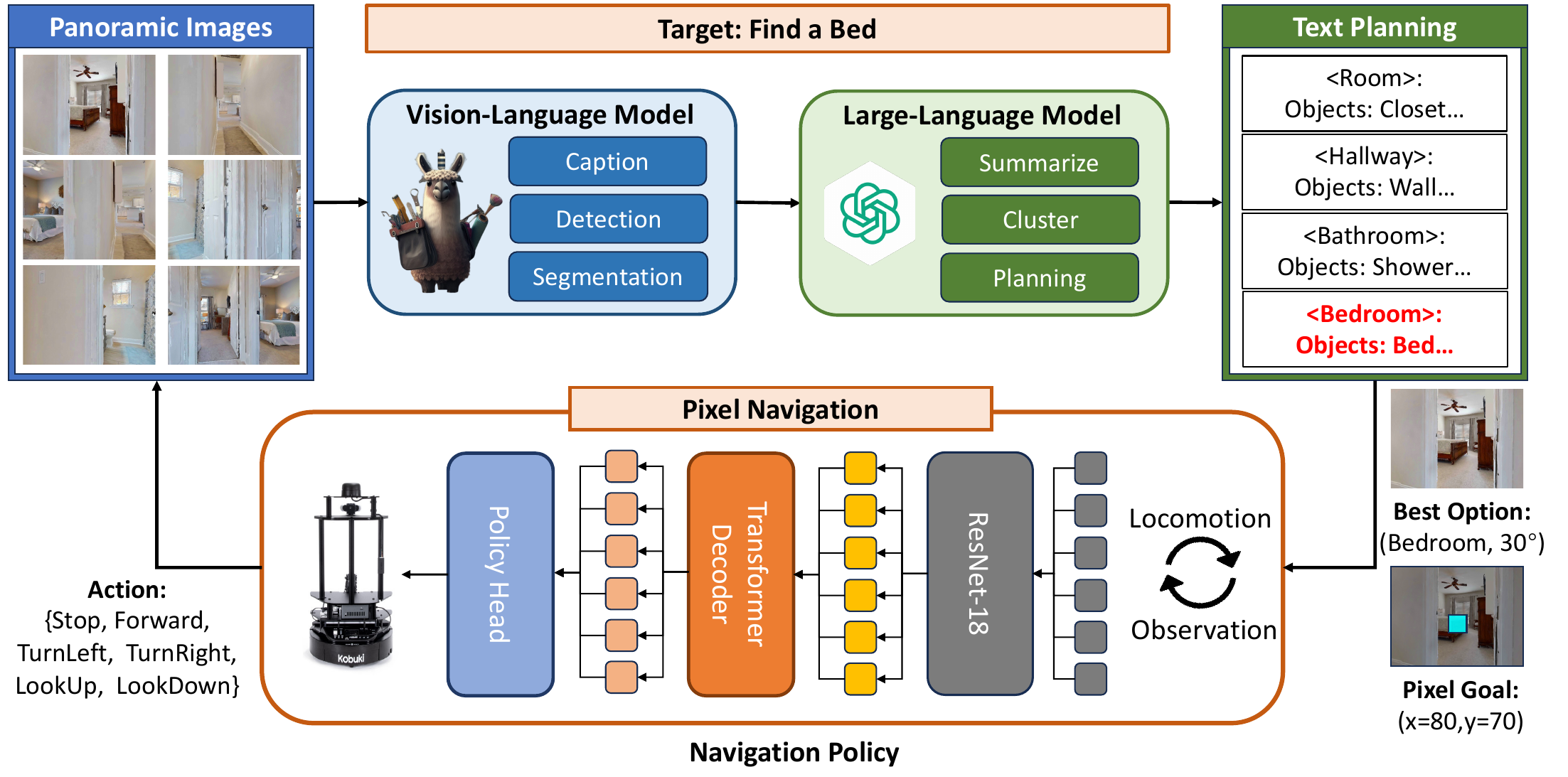}
\end{center}
\vspace{-0.35cm}
\caption{The pipeline of our RGB-centric strategy for zero-shot object navigation. In each cycle, the agent first gets the panoramic images of the surroundings. This can be finished by controlling the robot by turning around or installing multiple RGB cameras. And a vision-language model translates this visual data into a textual description. Utilizing a systematically crafted step-by-step prompt template, the LLM then strategizes the most optimal next step for the target location. Then, the target location will be indicated as a pixel and conveyed to the navigation policy. The PixNav policy will continuously receive observation images and perform actions until arriving at the goal area.}
\label{fig:architecture}
\vspace{-0.5cm}
\end{figure*}

\subsection{Large-Language Model for Visual Navigation}
For many visual navigation tasks, the agent is required to achieve efficient exploration by linking the goals and the observed semantic clues. Leveraging LLM's powerful commonsense reasoning ability~\cite{Brown2020LanguageMA, Driess2023PaLMEAE, OpenAI2023GPT4TR}, many works start to explore the ways to introduce the LLM's planning capability into visual navigation tasks.
For instance, L3MVN~\cite{Yu2023L3MVNLL} transcribes surrounding information of map boundaries into text, enabling the LLM to evaluate the scores of the frontiers for better exploration. ESC~\cite{zhou2023esc} further feeds both room-level and object-level information to LLM to select an exploration target. Similarly, LGX~\cite{dorbala2023can} also captions the image observation into text but uses LLM to make a search plan for an arbitrarily described object. StructNav~\cite{chen2023train} expands the scene memory with point clouds, scene graphs, and semantic occupancy maps, making more reasonable planning with LLM for object navigation.
NavGPT~\cite{Zhou2023NavGPTER} discovers a way to directly use the planning and reasoning capability of LLMs on the instruction-following navigation agent with prompt engineering. We follow the idea to introduce the LLM to promote the efficiency of an object navigation agent, but differ from those studies, we design a step-by-step prompt template and discover the potential of LLM in self-localization and structural memory organization, which can be further extended in other navigation tasks. However, our unique contribution lies in a crafted step-by-step prompt template, which could guide LLMs to pinpoint the agent's location and adeptly organize memory. And we believe these findings have potential implications for a broader range of navigation tasks.

\section{METHOD} 
\subsection{Path-Planning with Pixel Navigation}
In pursuit of a universally applicable navigation strategy, we introduce the pixel-guided navigation skill (PixNav) tailored for zero-shot object navigation tasks. The PixNav uses a pixel in the initial frame as the goal and executes primitive actions navigating towards the corresponding area indicated by the pixel. 
Our approach leverages an action space composed of six primitive actions: \{Stop, MoveAhead, TurnLeft, TurnRight, LookUp, LookDown\}. Notably, while our model currently operates in this discrete action space, PixNav has the potential to be extended to continuous settings. To help the model deal with the partial observability and tracking the pixel in the first frame, we use a transformer-based network with all history images $I_{0\sim T} = \{I_{0},I_{1}... I_{T}\}$ and the goal pixel $p_{g}=(x_{i},y_{i})$ as inputs. All the parameters are optimized from scratch and the architecture is summarized at the bottom part in Fig~\ref{fig:architecture}. 

\noindent \textbf{Training Data.} 
The large-scale dataset is essential for the robot learning~\cite{Brohan2022RT1RT}. Regarding the demonstration data collection, our PixNav skill offers marked advantages over the object-goal navigation approaches. The latter demands a unique object instance for each navigation trajectory, complicating the process through the necessity for object instance annotation prior to generating each trajectory. Such object instances are generally sparse within a single image which limits the potential collected training trajectories. Conversely, PixNav directly utilizes the vast potential of the pixel space, yielding much diverse navigation trajectories. The integration of the depth sensor further eases this approach, by converting each pixel to a 3D coordinate. Then, off-the-shelf path planners can be utilized, which is especially straightforward in simulation environments. In our work, we use the Habitat-Lab~\cite{habitat19iccv,szot2021habitat} with HM3D~\cite{Ramakrishnan2021HabitatMatterport3D} dataset to generate the training trajectories for PixNav. At the start of each episode, we randomly sample a pixel from the RGB observation and use the underlying path planner to interpolate the actions. In total, we generate a dataset with more than 100,000 trajectories to train the policy and we constrain the trajectory length within 64-time steps. The size of RGB observation in the trajectories is 160x120. 
\vspace{-0.1cm}

\noindent \textbf{Architecture Details.} To encode all the history images and the target pixel, we first convert the pixel goal $p_{g}=(x_{i},y_{i})$ into a mask image $M_{g}$ where the values within the bounding box $(x_{i}-\delta, y_{i}-\delta, x_{i}+\delta, y_{i}+\delta)$ are set to 1 and the rest are set to 0. Here, $\delta$ is a parameter used to control the masked area and in our work, we set it to 2. We then concatenate the mask image with the initial image in a channel-wise manner and employ a 4-channel ResNet18 to process the combined input $v_{g}=(I_{0}, M_{g})$, yielding a 768-dimensional goal token $\Phi_{\theta}(v_{g})$. For the RGB observation images, we use another 3-channel ResNet18 to obtain a sequence of 512-dimensional observation tokens $[\varphi_{\theta} (I_{0}),\varphi_{\theta} (I_{1}) ...\varphi_{\theta} (I_{T})]$. Both ResNet18 models are initialized randomly. To ensure the information of the goal pixel can be utilized by the transformer, we introduce a goal-fusion layer to generate a 256-dimensional feature from $\Phi_{\theta}(v_{g})$ and concatenate it with each observation tokens. Subsequently, a 4-layer transformer decoder is employed to integrate the global information across all tokens. Denote the output tokens as $[s_{0},s_{1}...s_{T}]$. These output tokens are fed into a policy layer to predict the optimal action $\pi_{\theta}(s_{i})$. As the robot performs the navigation task, the changes in viewpoint will cause the movement of the pixel goal in the subsequent frames. To help the model understand such transformation, we introduce a tracking layer to predict the corresponding pixel in every RGB observation $f_{\theta}(s_{i})$. Moreover, to aid the model in understanding the 3-D structure, we introduce a temporal distance layer to predict the remaining time steps ${d}_{\theta}(s_{i})$ to each the pixel goal. All the policy layer, tracking layer, and temporal distance layer are 1-layer MLP built upon output tokens from transformer decoder $[s_{0},s_{1}...s_{T}]$. The parameters of the entire network are 54.5M.

\noindent \textbf{Training Objective.} To train the entire network, we incorporate  three objectives. The first is the imitation learning objective for $\pi_{\theta}(s_{i})$, the second is MSE loss for the corresponding pixel prediction $f_{\theta}(s_{i})$,  and the third is MSE loss for temporal distance prediction ${d}_{\theta}(s_{i})$. Because the temporal distance can vary from 1 to 64, we scale the loss function for temporal distance by dividing it by 10. In summary, the objectives can be formulated as follows:
\begin{align}
        L_{il} &= \mathbb{E}_{\tau}[\hat{a}_{t}\log\pi_{\theta}(s_{t})] \\
        L_{distance} &= \mathbb{E}_{\tau}[(\hat{d_{t}}-d_{\theta}(s_{i}))^{2}] \\
        L_{track} &= \mathbb{E}_{\tau}[(\hat{p_{t}}-f_{\theta}(s_{i}))^{2}]
\end{align}

The $\hat{a_{t}},\hat{d_{t}},\hat{p_{t}}$ are the ground truth for best actions, temporal distance, and the corresponding pixel. Therefore, all the training processes are supervised learning.
The entire loss is written as follows:
\begin{equation}
    L = L_{il} + L_{distance} + L_{track}
\end{equation}

\subsection{Perception with Foundation Model}
\begin{figure}[h]
\begin{center}
    \includegraphics[width=1.0\linewidth]{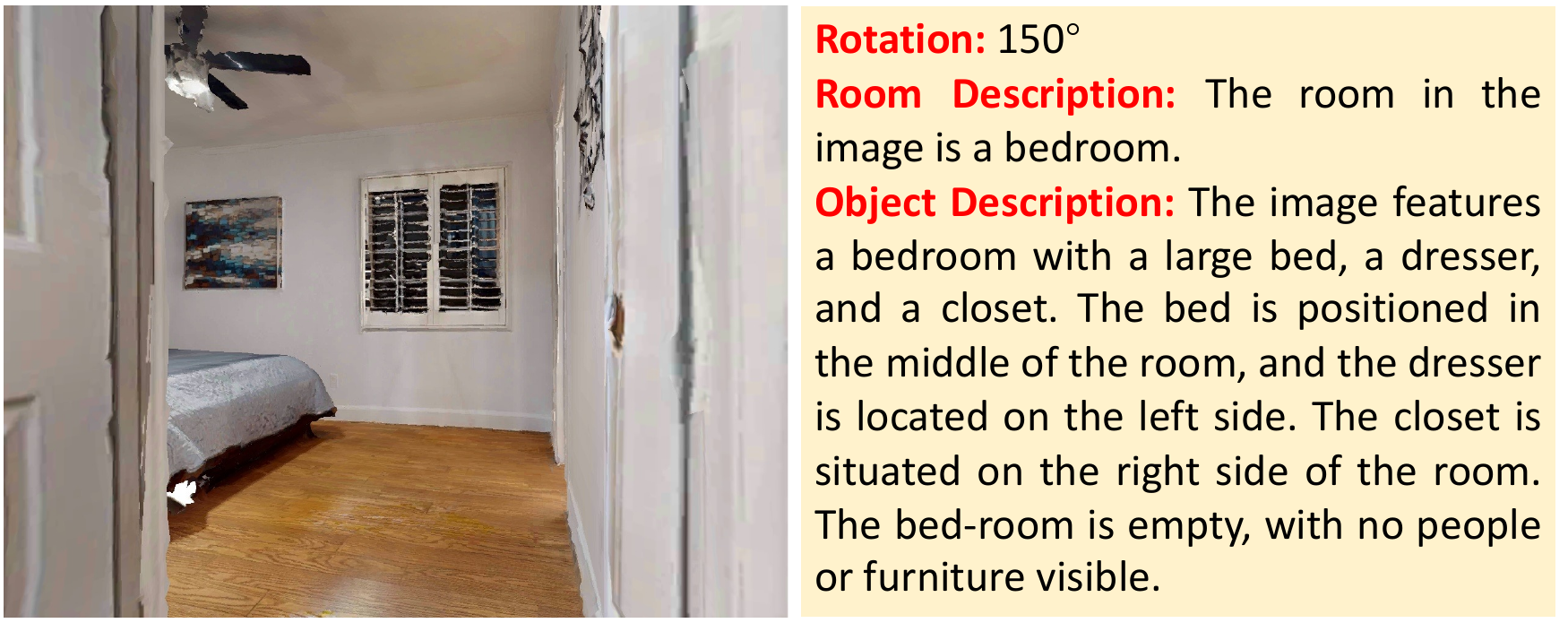}
\end{center}
\label{fig:vlm_example}
\vspace{-0.35cm}
\caption{An example of the VLM translation. The VLM is prompted with room-level and object-level queries. The robot performs a series of rotations, capturing six distinct images to ensure a comprehensive panoramic view.}
\vspace{-0.5cm}
\end{figure}

Despite the commonsense inference and reasoning capabilities contained in LLMs, navigation tasks generally operate with image-based observations. To bridge this modality gap, we employ the vision-language model, LLama-Adapter~\cite{Zhang2023LLaMAAdapterEF}, as a conduit for image summarization. Specifically, we prompt the VLM with two structured queries:  "Describe the room type in the image" and "Describe the objects with details in the image". Then, each image can be converted into a structured textual representation, as demonstrated in Fig~\ref{fig:vlm_example}. To make a comprehensive planning, a panoramic understanding is essential. To this end, the robot undergoes a series of rotations and captures six distinct images which are converted into six paragraphs as the input for LLM later. In order to allocate one pixel as the goal, we introduce two additional foundation models, namely GroundingDINO~\cite{Liu2023GroundingDM} and SAM~\cite{kirillov2023segment} to capture the find-grained segmentation masks. Once an object target is proposed by LLM, the center point of the object mask will be assigned as the pixel goal for the low-level navigation policy.

\subsection{Planning with Large-Language Model}
\begin{figure*}[t]
\begin{center}
    \includegraphics[width=0.90\linewidth]{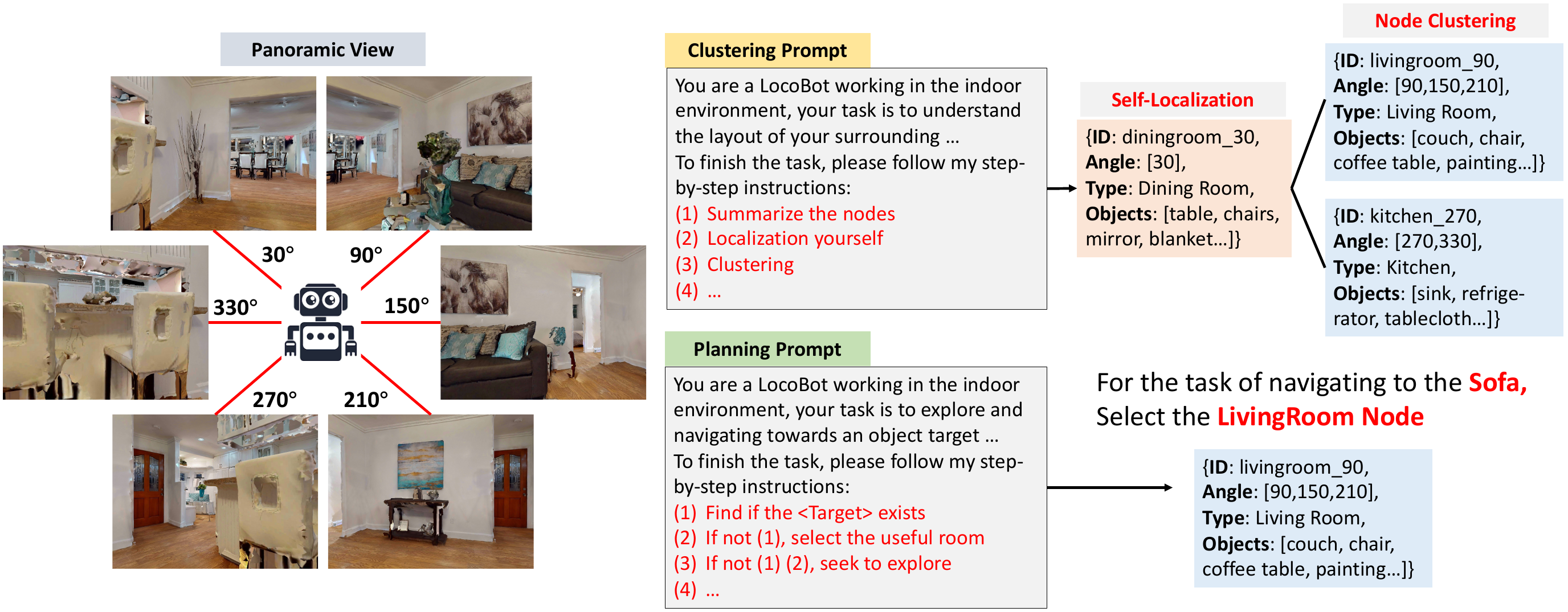}
\end{center}
\vspace{-0.35cm}
\caption{The step-by-step prompt template for LLM to make a reasonable plan for navigation. After translating the panoramic images into text, we first use the LLM to summarize the captions into highly structured data and ask LLM to estimate the robot's location according to the commonsense of indoor environments. With the robot location, we can cluster the data into room-level instances and thus the layout can be described as a graph. Based on this graph, we then ask the LLM to provide a plan considering both exploitation and exploration.}
\label{fig:planner}
\vspace{-0.5cm}
\end{figure*}

For the long-horizon object navigation task, we design an LLM-based planner to decide which direction is most promising for searching the target object. For example, a bed is usually placed in a bedroom, 
and the agent needs to traverse a corridor to reach the bedroom. To achieve this, we design step-to-step prompts to help the LLM understand an indoor navigation scenario. The entire planning procedure is divided into three stages: Summarize, Clustering, and Planning, as shown in Fig~\ref{fig:planner}. Firstly, as prior works show structured text is beneficial for LLM's planning and reasoning~\cite{rana2023sayplan,10161317}, we ask the LLM to summarize the VLM captions into JSON format which contains \{`ID', `Angle', `Type', `Objects'\}. Each JSON data encapsulates room-level information. However, the six surrounding images do not generally represent six unique room instances and several images can be grouped under a single room. To cluster the data, we use LLM to estimate the robot's current room and its position, and then build the inter-room connections. The same type of rooms with an edge connection will be merged into one. We also cluster the same type of rooms with adjacent angles. Then, a graph of the local layout in the house is generated by the LLM as shown at the top of Fig~\ref{fig:planner}. With this graph, the LLM can perform consistent exploration behaviors among rooms instead of roaming in one room. We design a planning prompt for the LLM to balance between exploration and exploitation for searching the target object. Our planner proposes the subgoal after periodic time steps or the low-level policy completes navigating towards the assigned goal. 

\section{EXPERIMENTAL RESULTS}

\subsection{Benchmarks and Metrics}
We employed the large-scale indoor dataset HM3D~\cite{Ramakrishnan2021HabitatMatterport3D} to evaluate our proposed method. The dataset contains 80 scenes for training and 20 scenes for validation. For the collection of PixNav demonstration trajectories, we utilized the 80 training scenes with the camera height 0.88m and horizontal field of view (HFOV) 79°. The evaluation for object navigation follows the settings in objectnav-challenge-2022~\cite{habitatchallenge2022}. For the object navigation task, we report the performance with respect to Success and SPL, which is a metric evaluating the navigation path efficiency. To evaluate the performance of PixNav, we randomly initialized the agent in 20 validation scenes and sampled a pixel goal in the first RGB frame. The PixNav evaluation is divided into two levels, one is 1-3m, and the other is 3-5m, which represents the underlying geodesic distance from the assigned pixel goals to the initial robot location. We further report the distance to goal (DTG), the geodesic distance from the location at the end of the trajectory to the pixel goal. For each level, we sample 500 episodes for evaluation. 
\vspace{-0cm}

\subsection{Baselines}
We consider both non-zero-shot object navigation methods and zero-shot object navigation methods to verify our idea. 
\begin{itemize}
    \item \noindent \textbf{Habitat-Web~\cite{Ramrakhya2022HabitatWebLE}.} A closed-set object navigation baseline with human demonstrations for imitation learning.
    \item \noindent \textbf{OVRL~\cite{Yadav2022OfflineVR}.} A closed-set object navigation baseline with the large-scale indoor dataset for representation learning.
    \item \noindent \textbf{ESC~\cite{zhou2023esc}} A map-based zero-shot object navigation baseline with LLM for frontier planning. 
    \item \noindent \textbf{ZSON~\cite{majumdar2022zson}} An RGB-based zero-shot object navigation baseline with CLIP~\cite{Radford2021LearningTV} as the image and object goal encoders.
\end{itemize}
\vspace{-0cm}

\subsection{Performance on Zero-Shot ObjectNav}
\begin{figure*}[h]
\begin{center}
    \includegraphics[width=1.0\linewidth]{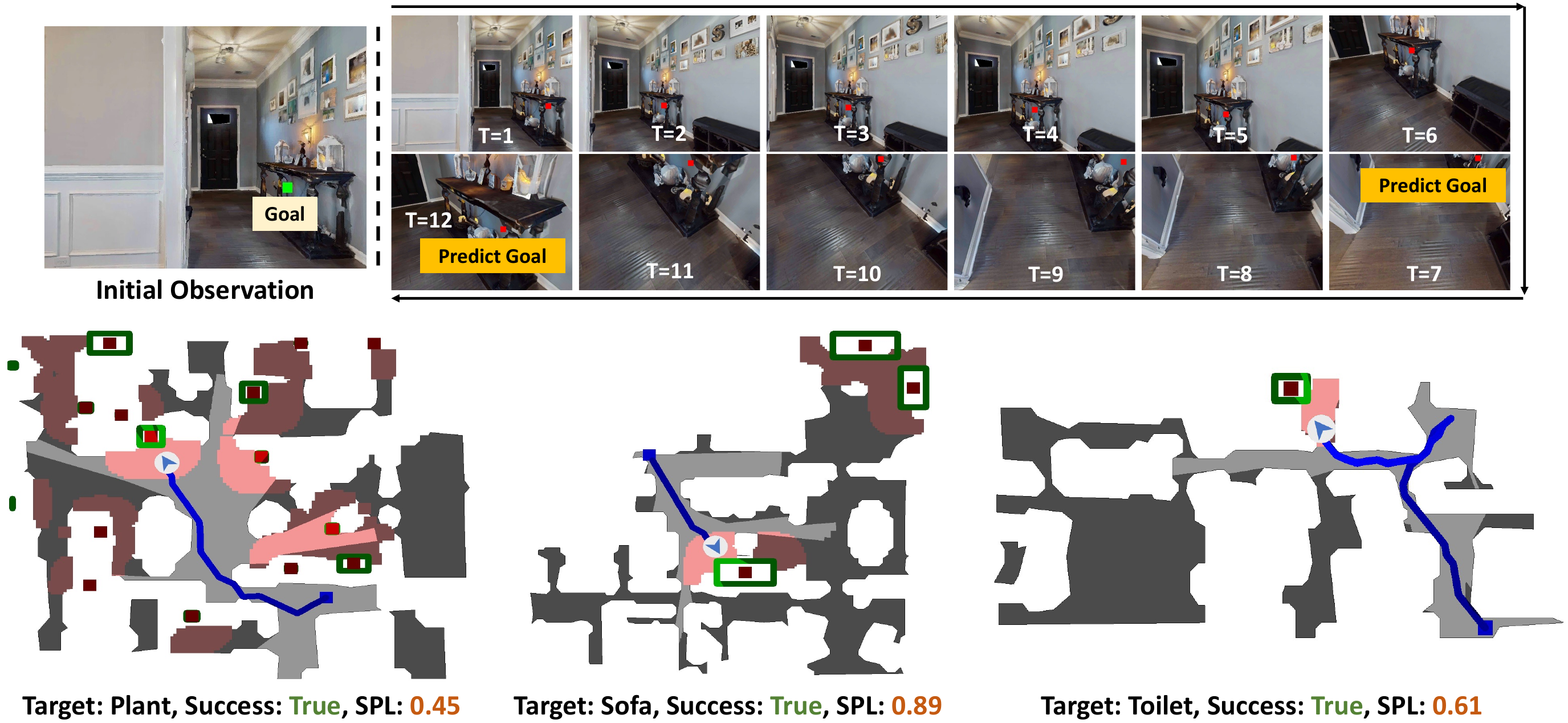}
\end{center}
\label{fig:trajectory}
\vspace{-0.5cm}
\caption{Trajectory visualization of both PixNav policy and long-horizon object navigation. The top part represents the first-person view of PixNav trajectory. Even though the robot's view changes constantly, our model can still predict the original goal and walk towards it. The bottom part represents the top-down view of the entire trajectory. Our LLM-based planner helps the robot achieve consistent and semantic exploration towards the target.}
\vspace{-0.3cm}
\end{figure*}

We compare our method with baseline results in Tab~\ref{tab:baseline} and example visualization of trajectories are shown in Fig~\ref{fig:trajectory}. Our approach achieves competitive performance even without depth and localization as inputs. Our PixNav can consistently track the initial pixel target and adaptively look down to avoid collision. The long-horizon trajectories show heuristic exploration behavior. For example, in the third figure, the agent can first go out of one room and enter another, then walk back to the corridor to search for a bathroom. It proves the effectiveness of our LLM-based planner. How to properly select the pixel as the goal for the PixNav is essential. We found that assigning the center pixel of the selected room 'floor' is a good strategy.

\begin{table}[ht]
  \centering
  \caption{Performance on object navigation tasks with HM3D dataset.}
  \begin{tabular}{ccccc}
      \toprule[1pt]
      \textbf{Setting} & \textbf{Mapless} & \textbf{Method} & \textbf{Success} & \textbf{SPL} \\
      \midrule[0.5pt]
      \multirow{2}{*}{\textit{Close-Set ObjectNav}} & \CheckmarkBold & Habitat-Web & 41.5 & 16.0 \\
      & \CheckmarkBold & OVRL & 62.0 & 26.8 \\ 
      \midrule[0.5pt]
      \multirow{3}{*}{\textit{Open-Set ObjectNav}} & 
      \CheckmarkBold & ZSON & 25.5 & 12.6 \\
      & \XSolidBrush & ESC & 39.2 & 22.3 \\
      & \CheckmarkBold & Ours & \textbf{37.9} & \textbf{20.5} \\
      \bottomrule[1pt]
  \end{tabular}
  \label{tab:baseline}
  \vspace{-0.5cm}
\end{table}

\subsection{Ablation Study on PixNav}
For real-world applications, robots may have different embodiment and camera settings. To become a general RGB-based path-planning module, the PixNav policy should be able to transfer to various camera configurations. Therefore, we adjusted the camera settings in the simulator with different heights and HFOV and re-evaluated the PixNav performance shown in Table~\ref{tab:camera-variation}. We found that even if we introduce a severe change in camera height (50\% relative change) and HFOV (25\% relative change), our policy can still achieve a high success rate in 1m-3m navigation tasks and degenerate little in most of the cases. This indicates that our PixNav can achieve zero-shot generalization to new camera settings and learn a universal representation for navigation. But what makes our policy robust? We conduct ablation studies on three components of our method, which are tracking prediction, temporal distance prediction, and goal fusion network. The performance is shown in Tab~\ref{tab:ablation}. Most of the components have a minor influence on the 1-3m scenarios, which shows the superiority of the transformer-based policy network architecture. However, tracking prediction and goal fusion significantly influence the 3-5m scenarios, emphasizing the need to explicitly remind the network not to overlook the inconspicuous pixel goal in order to achieve long-term navigation with PixNav.

\begin{table}[ht]
  \centering
  \caption{Performance on different variations of camera settings}
  \begin{tabular}{ccccccc}
      \toprule[1pt]
      \multirow{2}{*}{\textbf{Camera}} & \multicolumn{3}{c}{\textbf{1m-3m}} & \multicolumn{3}{c}{\textbf{3m-5m}} \\
      \cmidrule(lr){2-7}
      & \textbf{SR $\uparrow$} & \textbf{SPL $\uparrow$} & \textbf{DTS $\downarrow$} & \textbf{SR $\uparrow$} & \textbf{SPL $\uparrow$} & \textbf{DTS $\downarrow$} \\
      \midrule[0.5pt]
      (0.88, 79°) & 0.88 & 0.88 & 0.54 & 0.42 & 0.41 & 1.61 \\
      \midrule[0.5pt]
      (0.48, 79°) & 0.79 & 0.76 & 0.61 & 0.26 & 0.25 & 1.95\\
      \midrule[0.5pt]
      (1.28, 79°) & 0.78 & 0.75 & 0.86 & 0.33 & 0.32 & 1.83 \\
      \midrule[0.5pt]
      (0.88, 60°) & 0.71 & 0.70 & 0.82 & 0.14 & 0.13 & 2.32 \\
      \midrule[0.5pt]
      (0.88, 100°) & 0.80 & 0.78 & 0.69 & 0.33 & 0.32 & 1.88 \\
      \bottomrule[1pt]
  \end{tabular}
  \label{tab:camera-variation}
  \vspace{-0.2cm}
\end{table}

\begin{table}[ht]
  \centering
  \caption{The ablation study of the PixNav in HM3D dataset}
  \begin{tabular}{ccccccc}
      \toprule[1pt]
      \multirow{2}{*}{\textbf{Ablation}} & \multicolumn{3}{c}{\textbf{1m-3m}} & \multicolumn{3}{c}{\textbf{3m-5m}} \\
      \cmidrule(lr){2-7}
      & \textbf{SR $\uparrow$} & \textbf{SPL $\uparrow$} & \textbf{DTS $\downarrow$} & \textbf{SR $\uparrow$} & \textbf{SPL $\uparrow$} & \textbf{DTS $\downarrow$} \\
      \midrule[0.5pt]
      Ours & \textbf{0.88} & \textbf{0.88} & \textbf{0.54} & \textbf{0.42} & \textbf{0.41} & \textbf{1.61} \\
      \midrule[0.5pt]
      w/o tracking & 0.82 & 0.81 & 0.63 & 0.33 & 0.31 & 1.75 \\
      \midrule[0.5pt]
      w/o distance & 0.86 & 0.85 & 0.60 & 0.38 & 0.35 & 1.69 \\
      \midrule[0.5pt]
      w/o fusion & 0.84 & 0.84 & 0.63 & 0.34 & 0.33 & 1.91 \\
      \bottomrule[1pt]
  \end{tabular}
  \label{tab:ablation}
  \vspace{-0.5cm}
\end{table}

\subsection{Analysis of LLM Planner}
To achieve long-horizon navigation tasks, the LLM planner is also an essential part. Traditional map-based approaches prove that mappping and localization are valuable skills for navigation tasks. If the LLM  also possesses these skills, the potential of our LLM-based RGB policy is promising for navigation field.
To probe on the abilities of LLM, we implement a quantitive analytic to verify whether the LLM owns such spatial-awareness of the indoor environments. The experiments contain two skills evaluations: Room Localization and Room Clustering. We ask 3 humans to see the panoramic images of the robot's location and answer two questions: "Which room is the robot in?" and "Describe the number and type of the rooms in the surroundings." We compare the answers from LLM and humans for 100 episodes. The accuracy of the two skills is shown in Tab~\ref{tab:LLM-Ability}. 

\begin{table}[ht]
    \centering
    \caption{The analysis of LLM Abilities in Localization and Mapping}
    \begin{tabular}{cc}
         \toprule[1pt]
         \textbf{LLM Ability} & \textbf{Human Eval (\%)} \\
         \midrule[0.5pt]
         Room Localization & 68.4 \\
         \midrule[0.5pt]
         Room Clustering & 47.1 \\
         \bottomrule[1pt]
    \end{tabular}
    \label{tab:LLM-Ability}
    \vspace{-0.5cm}
\end{table}
\subsection{Sim2Real Transfer}
The real-world experiments are conducted with an iRobot-Create3 Mobile base equipped with an OAK-D camera. However, we only use the RGB observations from the camera for the experiments. The action configuration of the robot follows the discrete action space setting, which contains basic actions \{Stop, MoveAhead 0.25m, TurnLeft 30°, TurnRight 30°\}. We conduct experiments to verify the function of both the PixNav policy as well as the LLM-planner. Please refer to our website for the details of real-world experiments.

\section{CONCLUSION}
In this work, we take a significant stride towards developing a pure RGB-based solution for visual navigation tasks through the utilization of a pixel-guided navigation skill. Our PixNav policy demonstrates versatility in navigating various objects, and it can effortlessly be integrated with advanced foundation models to tackle long-horizon navigation tasks. Moreover, PixNav exhibits generalization capabilities across different camera settings, showcasing its potential to serve as a versatile RGB-based path-planning policy. However, the current performance of PixNav in long-term path-planning is not satisfactory. Therefore, in future research, we will investigate the potential benefits of large-scale and diverse navigation datasets. Additionally, the design of the navigation-grounded vision-language model as a planner would also bring improvements in the object-navigation domain.

\clearpage
{
\bibliographystyle{IEEEtran}
\bibliography{IEEEabrv,reference}

\begin{thebibliography}{10}
\providecommand{\url}[1]{#1}
\csname url@samestyle\endcsname
\providecommand{\newblock}{\relax}
\providecommand{\bibinfo}[2]{#2}
\providecommand{\BIBentrySTDinterwordspacing}{\spaceskip=0pt\relax}
\providecommand{\BIBentryALTinterwordstretchfactor}{4}
\providecommand{\BIBentryALTinterwordspacing}{\spaceskip=\fontdimen2\font plus
\BIBentryALTinterwordstretchfactor\fontdimen3\font minus
  \fontdimen4\font\relax}
\providecommand{\BIBforeignlanguage}[2]{{%
\expandafter\ifx\csname l@#1\endcsname\relax
\typeout{** WARNING: IEEEtran.bst: No hyphenation pattern has been}%
\typeout{** loaded for the language `#1'. Using the pattern for}%
\typeout{** the default language instead.}%
\else
\language=\csname l@#1\endcsname
\fi
#2}}
\providecommand{\BIBdecl}{\relax}
\BIBdecl

\bibitem{Liu2023GroundingDM}
\BIBentryALTinterwordspacing
S.~Liu, Z.~Zeng, T.~Ren, F.~Li, H.~Zhang, J.~Yang, C.~yue Li, J.~Yang, H.~Su,
  J.-J. Zhu, and L.~Zhang, ``Grounding dino: Marrying dino with grounded
  pre-training for open-set object detection,'' \emph{ArXiv}, vol.
  abs/2303.05499, 2023. [Online]. Available:
  \url{https://api.semanticscholar.org/CorpusID:257427307}
\BIBentrySTDinterwordspacing

\bibitem{Minderer2022SimpleOO}
\BIBentryALTinterwordspacing
M.~Minderer, A.~A. Gritsenko, A.~Stone, M.~Neumann, D.~Weissenborn,
  A.~Dosovitskiy, A.~Mahendran, A.~Arnab, M.~Dehghani, Z.~Shen, X.~Wang,
  X.~Zhai, T.~Kipf, and N.~Houlsby, ``Simple open-vocabulary object detection
  with vision transformers,'' \emph{ArXiv}, vol. abs/2205.06230, 2022.
  [Online]. Available: \url{https://api.semanticscholar.org/CorpusID:248721818}
\BIBentrySTDinterwordspacing

\bibitem{kirillov2023segment}
A.~Kirillov, E.~Mintun, N.~Ravi, H.~Mao, C.~Rolland, L.~Gustafson, T.~Xiao,
  S.~Whitehead, A.~C. Berg, W.-Y. Lo \emph{et~al.}, ``Segment anything,''
  \emph{arXiv preprint arXiv:2304.02643}, 2023.

\bibitem{Li2022LanguagedrivenSS}
\BIBentryALTinterwordspacing
B.~Li, K.~Q. Weinberger, S.~J. Belongie, V.~Koltun, and R.~Ranftl,
  ``Language-driven semantic segmentation,'' \emph{ArXiv}, vol. abs/2201.03546,
  2022. [Online]. Available:
  \url{https://api.semanticscholar.org/CorpusID:245836975}
\BIBentrySTDinterwordspacing

\bibitem{zhang2023llama}
R.~Zhang, J.~Han, A.~Zhou, X.~Hu, S.~Yan, P.~Lu, H.~Li, P.~Gao, and Y.~Qiao,
  ``Llama-adapter: Efficient fine-tuning of language models with zero-init
  attention,'' \emph{arXiv preprint arXiv:2303.16199}, 2023.

\bibitem{Dai2023InstructBLIPTG}
\BIBentryALTinterwordspacing
W.~Dai, J.~Li, D.~Li, A.~M.~H. Tiong, J.~Zhao, W.~Wang, B.~Li, P.~Fung, and
  S.~C.~H. Hoi, ``Instructblip: Towards general-purpose vision-language models
  with instruction tuning,'' \emph{ArXiv}, vol. abs/2305.06500, 2023. [Online].
  Available: \url{https://api.semanticscholar.org/CorpusID:258615266}
\BIBentrySTDinterwordspacing

\bibitem{Gadre2022CLIPOW}
S.~Y. Gadre, M.~Wortsman, G.~Ilharco, L.~Schmidt, and S.~Song, ``Clip on
  wheels: Zero-shot object navigation as object localization and exploration,''
  \emph{ArXiv}, vol. abs/2203.10421, 2022.

\bibitem{zhou2023esc}
K.~Zhou, K.~Zheng, C.~Pryor, Y.~Shen, H.~Jin, L.~Getoor, and X.~E. Wang, ``Esc:
  Exploration with soft commonsense constraints for zero-shot object
  navigation,'' \emph{arXiv preprint arXiv:2301.13166}, 2023.

\bibitem{Yu2023L3MVNLL}
B.~Yu, H.~Kasaei, and M.~Cao, ``L3mvn: Leveraging large language models for
  visual target navigation,'' \emph{ArXiv}, vol. abs/2304.05501, 2023.

\bibitem{dorbala2023can}
V.~S. Dorbala, J.~F. Mullen~Jr, and D.~Manocha, ``Can an embodied agent find
  your" cat-shaped mug"? llm-based zero-shot object navigation,'' \emph{arXiv
  preprint arXiv:2303.03480}, 2023.

\bibitem{chen2023train}
J.~Chen, G.~Li, S.~Kumar, B.~Ghanem, and F.~Yu, ``How to not train your dragon:
  Training-free embodied object goal navigation with semantic frontiers,''
  2023.

\bibitem{khandelwal2022:embodied-clip}
A.~Khandelwal, L.~Weihs, R.~Mottaghi, and A.~Kembhavi, ``Simple but effective:
  Clip embeddings for embodied ai,'' in \emph{Proceedings of the IEEE/CVF
  Conference on Computer Vision and Pattern Recognition (CVPR)}, June 2022.

\bibitem{chaplot2020object}
D.~S. Chaplot, D.~Gandhi, A.~Gupta, and R.~Salakhutdinov, ``Object goal
  navigation using goal-oriented semantic exploration,'' in \emph{In Neural
  Information Processing Systems (NeurIPS)}, 2020.

\bibitem{Brohan2022RT1RT}
\BIBentryALTinterwordspacing
A.~Brohan, N.~Brown, J.~Carbajal, Y.~Chebotar, J.~Dabis, C.~Finn,
  K.~Gopalakrishnan, K.~Hausman, A.~Herzog, J.~Hsu, J.~Ibarz, B.~Ichter,
  A.~Irpan, T.~Jackson, S.~Jesmonth, N.~J. Joshi, R.~C. Julian, D.~Kalashnikov,
  Y.~Kuang, I.~Leal, K.-H. Lee, S.~Levine, Y.~Lu, U.~Malla, D.~Manjunath,
  I.~Mordatch, O.~Nachum, C.~Parada, J.~Peralta, E.~Perez, K.~Pertsch,
  J.~Quiambao, K.~Rao, M.~S. Ryoo, G.~Salazar, P.~R. Sanketi, K.~Sayed,
  J.~Singh, S.~A. Sontakke, A.~Stone, C.~Tan, H.~Tran, V.~Vanhoucke, S.~Vega,
  Q.~H. Vuong, F.~Xia, T.~Xiao, P.~Xu, S.~Xu, T.~Yu, and B.~Zitkovich, ``Rt-1:
  Robotics transformer for real-world control at scale,'' \emph{ArXiv}, vol.
  abs/2212.06817, 2022. [Online]. Available:
  \url{https://api.semanticscholar.org/CorpusID:254591260}
\BIBentrySTDinterwordspacing

\bibitem{Walke2023BridgeDataVA}
\BIBentryALTinterwordspacing
H.~Walke, K.~Black, A.~Lee, M.~J. Kim, M.~Du, C.~Zheng, T.~Zhao,
  P.~Hansen-Estruch, Q.~H. Vuong, A.~W. He, V.~Myers, K.~Fang, C.~Finn, and
  S.~Levine, ``Bridgedata v2: A dataset for robot learning at scale,''
  \emph{ArXiv}, vol. abs/2308.12952, 2023. [Online]. Available:
  \url{https://api.semanticscholar.org/CorpusID:261100981}
\BIBentrySTDinterwordspacing

\bibitem{rramrakhya2022}
R.~Ramrakhya, E.~Undersander, D.~Batra, and A.~Das, ``Habitat-web: Learning
  embodied object-search strategies from human demonstrations at scale,'' in
  \emph{CVPR}, 2022.

\bibitem{ramrakhya2023pirlnav}
R.~Ramrakhya, D.~Batra, E.~Wijmans, and A.~Das, ``Pirlnav: Pretraining with
  imitation and rl finetuning for objectnav,'' in \emph{CVPR}, 2023.

\bibitem{yadav2023offline}
\BIBentryALTinterwordspacing
K.~Yadav, R.~Ramrakhya, A.~Majumdar, V.-P. Berges, S.~Kuhar, D.~Batra,
  A.~Baevski, and O.~Maksymets, ``Offline visual representation learning for
  embodied navigation,'' in \emph{Workshop on Reincarnating Reinforcement
  Learning at ICLR 2023}, 2023. [Online]. Available:
  \url{https://openreview.net/forum?id=Spfbts_vNY}
\BIBentrySTDinterwordspacing

\bibitem{Zhu2022NavigatingTO}
M.~Zhu, B.~Zhao, and T.~Kong, ``Navigating to objects in unseen environments by
  distance prediction,'' \emph{2022 IEEE/RSJ International Conference on
  Intelligent Robots and Systems (IROS)}, pp. 10\,571--10\,578, 2022.

\bibitem{majumdar2022zson}
A.~Majumdar, G.~Aggarwal, B.~Devnani, J.~Hoffman, and D.~Batra, ``Zson:
  Zero-shot object-goal navigation using multimodal goal embeddings,'' in
  \emph{Neural Information Processing Systems (NeurIPS)}, 2022.

\bibitem{Radford2021LearningTV}
A.~Radford, J.~W. Kim, C.~Hallacy, A.~Ramesh, G.~Goh, S.~Agarwal, G.~Sastry,
  A.~Askell, P.~Mishkin, J.~Clark, G.~Krueger, and I.~Sutskever, ``Learning
  transferable visual models from natural language supervision,'' in
  \emph{International Conference on Machine Learning}, 2021.

\bibitem{Zhao2022ZeroShotOG}
Q.~Zhao, L.~Zhang, B.~He, H.~Qiao, and Z.~yong Liu, ``Zero-shot object goal
  visual navigation,'' \emph{2023 IEEE International Conference on Robotics and
  Automation (ICRA)}, pp. 2025--2031, 2022.

\bibitem{Huang2022VisualLM}
C.~Huang, O.~Mees, A.~Zeng, and W.~Burgard, ``Visual language maps for robot
  navigation,'' \emph{2023 IEEE International Conference on Robotics and
  Automation (ICRA)}, pp. 10\,608--10\,615, 2022.

\bibitem{Chen2022OpenvocabularyQS}
B.~Chen, F.~Xia, B.~Ichter, K.~Rao, K.~Gopalakrishnan, M.~S. Ryoo, A.~Stone,
  and D.~Kappler, ``Open-vocabulary queryable scene representations for real
  world planning,'' \emph{2023 IEEE International Conference on Robotics and
  Automation (ICRA)}, pp. 11\,509--11\,522, 2022.

\bibitem{Zhang2023RecognizeAA}
\BIBentryALTinterwordspacing
Y.~Zhang, X.~Huang, J.~Ma, Z.~Li, Z.~Luo, Y.~Xie, Y.~Qin, T.~Luo, Y.~Li,
  S.~Liu, Y.~Guo, and L.~Zhang, ``Recognize anything: A strong image tagging
  model,'' \emph{ArXiv}, vol. abs/2306.03514, 2023. [Online]. Available:
  \url{https://api.semanticscholar.org/CorpusID:259089333}
\BIBentrySTDinterwordspacing

\bibitem{Touvron2023LLaMAOA}
\BIBentryALTinterwordspacing
H.~Touvron, T.~Lavril, G.~Izacard, X.~Martinet, M.-A. Lachaux, T.~Lacroix,
  B.~Rozi{\`e}re, N.~Goyal, E.~Hambro, F.~Azhar, A.~Rodriguez, A.~Joulin,
  E.~Grave, and G.~Lample, ``Llama: Open and efficient foundation language
  models,'' \emph{ArXiv}, vol. abs/2302.13971, 2023. [Online]. Available:
  \url{https://api.semanticscholar.org/CorpusID:257219404}
\BIBentrySTDinterwordspacing

\bibitem{Brown2020LanguageMA}
T.~B. Brown, B.~Mann, N.~Ryder, M.~Subbiah, J.~Kaplan, P.~Dhariwal,
  A.~Neelakantan, P.~Shyam, G.~Sastry, A.~Askell, S.~Agarwal, A.~Herbert-Voss,
  G.~Krueger, T.~J. Henighan, R.~Child, A.~Ramesh, D.~M. Ziegler, J.~Wu,
  C.~Winter, C.~Hesse, M.~Chen, E.~Sigler, M.~Litwin, S.~Gray, B.~Chess,
  J.~Clark, C.~Berner, S.~McCandlish, A.~Radford, I.~Sutskever, and D.~Amodei,
  ``Language models are few-shot learners,'' \emph{ArXiv}, vol. abs/2005.14165,
  2020.

\bibitem{Huang2022LanguageMA}
\BIBentryALTinterwordspacing
W.~Huang, P.~Abbeel, D.~Pathak, and I.~Mordatch, ``Language models as zero-shot
  planners: Extracting actionable knowledge for embodied agents,''
  \emph{ArXiv}, vol. abs/2201.07207, 2022. [Online]. Available:
  \url{https://api.semanticscholar.org/CorpusID:246035276}
\BIBentrySTDinterwordspacing

\bibitem{10161317}
I.~Singh, V.~Blukis, A.~Mousavian, A.~Goyal, D.~Xu, J.~Tremblay, D.~Fox,
  J.~Thomason, and A.~Garg, ``Progprompt: Generating situated robot task plans
  using large language models,'' in \emph{2023 IEEE International Conference on
  Robotics and Automation (ICRA)}, 2023, pp. 11\,523--11\,530.

\bibitem{rana2023sayplan}
K.~Rana, J.~Haviland, S.~Garg, J.~Abou-Chakra, I.~Reid, and N.~Suenderhauf,
  ``Sayplan: Grounding large language models using 3d scene graphs for scalable
  task planning,'' \emph{arXiv preprint arXiv:2307.06135}, 2023.

\bibitem{Huang2023Instruct2ActMM}
\BIBentryALTinterwordspacing
S.~Huang, Z.~Jiang, H.-W. Dong, Y.~J. Qiao, P.~Gao, and H.~Li, ``Instruct2act:
  Mapping multi-modality instructions to robotic actions with large language
  model,'' \emph{ArXiv}, vol. abs/2305.11176, 2023. [Online]. Available:
  \url{https://api.semanticscholar.org/CorpusID:258762636}
\BIBentrySTDinterwordspacing

\bibitem{Zhu2023MiniGPT4EV}
\BIBentryALTinterwordspacing
D.~Zhu, J.~Chen, X.~Shen, X.~Li, and M.~Elhoseiny, ``Minigpt-4: Enhancing
  vision-language understanding with advanced large language models,''
  \emph{ArXiv}, vol. abs/2304.10592, 2023. [Online]. Available:
  \url{https://api.semanticscholar.org/CorpusID:258291930}
\BIBentrySTDinterwordspacing

\bibitem{Zhang2023LLaMAAdapterEF}
\BIBentryALTinterwordspacing
R.~Zhang, J.~Han, A.~Zhou, X.~Hu, S.~Yan, P.~Lu, H.~Li, P.~Gao, and Y.~J. Qiao,
  ``Llama-adapter: Efficient fine-tuning of language models with zero-init
  attention,'' \emph{ArXiv}, vol. abs/2303.16199, 2023. [Online]. Available:
  \url{https://api.semanticscholar.org/CorpusID:257771811}
\BIBentrySTDinterwordspacing

\bibitem{Driess2023PaLMEAE}
D.~Driess, F.~Xia, M.~S.~M. Sajjadi, C.~Lynch, A.~Chowdhery, B.~Ichter,
  A.~Wahid, J.~Tompson, Q.~H. Vuong, T.~Yu, W.~Huang, Y.~Chebotar, P.~Sermanet,
  D.~Duckworth, S.~Levine, V.~Vanhoucke, K.~Hausman, M.~Toussaint, K.~Greff,
  A.~Zeng, I.~Mordatch, and P.~R. Florence, ``Palm-e: An embodied multimodal
  language model,'' \emph{ArXiv}, vol. abs/2303.03378, 2023.

\bibitem{OpenAI2023GPT4TR}
OpenAI, ``Gpt-4 technical report,'' \emph{ArXiv}, vol. abs/2303.08774, 2023.

\bibitem{Zhou2023NavGPTER}
G.~Zhou, Y.~Hong, and Q.~Wu, ``Navgpt: Explicit reasoning in
  vision-and-language navigation with large language models,'' \emph{ArXiv},
  vol. abs/2305.16986, 2023.

\bibitem{habitat19iccv}
M.~Savva, A.~Kadian, O.~Maksymets, Y.~Zhao, E.~Wijmans, B.~Jain, J.~Straub,
  J.~Liu, V.~Koltun, J.~Malik, D.~Parikh, and D.~Batra, ``Habitat: {A}
  {P}latform for {E}mbodied {AI} {R}esearch,'' in \emph{Proceedings of the
  IEEE/CVF International Conference on Computer Vision (ICCV)}, 2019.

\bibitem{szot2021habitat}
A.~Szot, A.~Clegg, E.~Undersander, E.~Wijmans, Y.~Zhao, J.~Turner, N.~Maestre,
  M.~Mukadam, D.~Chaplot, O.~Maksymets, A.~Gokaslan, V.~Vondrus, S.~Dharur,
  F.~Meier, W.~Galuba, A.~Chang, Z.~Kira, V.~Koltun, J.~Malik, M.~Savva, and
  D.~Batra, ``Habitat 2.0: Training home assistants to rearrange their
  habitat,'' in \emph{Advances in Neural Information Processing Systems
  (NeurIPS)}, 2021.

\bibitem{Ramakrishnan2021HabitatMatterport3D}
\BIBentryALTinterwordspacing
S.~K. Ramakrishnan, A.~Gokaslan, E.~Wijmans, O.~Maksymets, A.~Clegg, J.~Turner,
  E.~Undersander, W.~Galuba, A.~Westbury, A.~X. Chang, M.~Savva, Y.~Zhao, and
  D.~Batra, ``Habitat-matterport 3d dataset (hm3d): 1000 large-scale 3d
  environments for embodied ai,'' \emph{ArXiv}, vol. abs/2109.08238, 2021.
  [Online]. Available: \url{https://api.semanticscholar.org/CorpusID:237563216}
\BIBentrySTDinterwordspacing

\bibitem{habitatchallenge2022}
K.~Yadav, J.~Krantz, R.~Ramrakhya, S.~K. Ramakrishnan, J.~Yang, A.~Wang,
  J.~Turner, A.~Gokaslan, V.-P. Berges, R.~Mootaghi, O.~Maksymets, A.~X. Chang,
  M.~Savva, A.~Clegg, D.~S. Chaplot, and D.~Batra,
  \url{https://aihabitat.org/challenge/2022/}, 2022.

\bibitem{Ramrakhya2022HabitatWebLE}
\BIBentryALTinterwordspacing
R.~Ramrakhya, E.~Undersander, D.~Batra, and A.~Das, ``Habitat-web: Learning
  embodied object-search strategies from human demonstrations at scale,''
  \emph{2022 IEEE/CVF Conference on Computer Vision and Pattern Recognition
  (CVPR)}, pp. 5163--5173, 2022. [Online]. Available:
  \url{https://api.semanticscholar.org/CorpusID:248006501}
\BIBentrySTDinterwordspacing

\bibitem{Yadav2022OfflineVR}
\BIBentryALTinterwordspacing
K.~Yadav, R.~Ramrakhya, A.~Majumdar, V.-P. Berges, S.~Kuhar, D.~Batra,
  A.~Baevski, and O.~Maksymets, ``Offline visual representation learning for
  embodied navigation,'' \emph{ArXiv}, vol. abs/2204.13226, 2022. [Online].
  Available: \url{https://api.semanticscholar.org/CorpusID:248426942}
\BIBentrySTDinterwordspacing

\end{thebibliography}
}
\end{document}